\RequirePackage{fix-cm}
\documentclass[twocolumn]{svjour3}          
\smartqed  
\usepackage{graphicx}
\usepackage{amsmath}
\usepackage{amssymb}
\usepackage{dsfont}

\usepackage{algpseudocode}
\usepackage{algpascal}
\usepackage{algorithm2e}

\usepackage{xcolor}

\newcommand{\ie}{\textit{i}.\textit{e}.}
\newcommand{\eg}{\textit{e}.\textit{g}.}

 \journalname{International Journal of Computer Vision}
\begin{document}

\title{Adding Knowledge to Unsupervised Algorithms for the Recognition of Intent
}


\author{Stuart Synakowski         \and
        Qianli Feng   \and
        Aleix Martinez 
}


\institute{S. Synakowski* \at
              Dept. Electrical and Computer Eng.\\
              The Ohio State University \\
              Columbus, OH 43210\\
              \email{synakowski.1@osu.edu} 
              \\
           \and
           Qianli Feng* \at
              Dept. Electrical and Computer Eng.\\
              The Ohio State University \\
              Columbus, OH 43210\\
              \email{feng.559@osu.edu} 
              \\
              \and
           Aleix Martinez\at
              Dept. Electrical and Computer Eng.\\
              The Ohio State University \\
              Columbus, OH 43210\\
              \email{martinez.158@osu.edu}    
              \\
              \and
              *equal contribution.\\
              This is a pre-print of an article published in International Journal of Computer Vision. The final authenticated version is available online at: https://doi.org/10.1007/s11263-020-01404-0
}

\date{Received: date / Accepted: date}
\maketitle

\begin{abstract}
Computer vision algorithms performance are near or superior to humans in the visual problems including object recognition (especially those of fine-grained categories), segmentation, and 3D object reconstruction from 2D views. Humans are, however, capable of higher-level image analyses. A clear example, involving theory of mind, is our ability to determine whether a perceived behavior or action was performed intentionally or not. In this paper, we derive an algorithm that can infer whether the behavior of an agent in a scene is intentional or unintentional based on its 3D kinematics, using the knowledge of self-propelled motion, Newtonian motion and their relationship. We show how the addition of this basic knowledge leads to a simple, unsupervised algorithm. To test the derived algorithm, we constructed three dedicated datasets from abstract geometric animation to realistic videos of agents performing intentional and non-intentional actions. Experiments on these datasets show that our algorithm can recognize whether an action is intentional or not, even without training data. The performance is comparable to various supervised baselines quantitatively, with sensible intentionality segmentation qualitatively.  

\end{abstract}

\section{Introduction}
\label{intro}
To solve high-level computer vision problems, like object recognition and action understanding, researchers and practitioners typically use very large datasets of manually labeled data to train a machine learning algorithm. These algorithms are typically used to discriminate between different categories, e.g., cars versus bikes, or running versus walking \cite{chen2019hybrid,yeung2018every}. Ideally, one would want to be able to design systems that can perform high-level task like these without the need of any manually annotated data. 

One way to derive such unsupervised computer vision algorithms is to incorporate knowledge into the system. Here, we derive one such approach and use it to recognize intentional and non-intentional actions. We use Aristotle's definition of intent as something deliberate, chosen before the start of the action \cite{aristotle1926art}. This definition is also included in Cartesian dualism, where Descartes differentiated conscious, intentional actions from reflexes caused by external stimuli \cite{descartes1960meditations}. 

To successfully classify a perceived action as intentional or unintentional, we need to carefully evaluate each segment of the video sequence displaying it. To clarify, consider the following example. A person is walking down a hall and after a few seconds slips and falls to the ground (maybe the floor is wet). Here, we would say that the person was intentionally walking down the hall, but that he unintentionally slipped and fell. Afterwards, he intentionally stood up and continued walking. Compare this to the case where the person does not slip but is instead pushed to the ground by someone else. In this case, we say that all segments in the scene are performed intentionally (since the fall is the result of the intentional push). Our goal is to derive an algorithm that can correctly and fully automatically annotate each segment of a video sequence as showing an intentional or a non-intentional action.

As mentioned above, to solve this problem, one could manually annotate a large number of video segments showing intentional and non-intentional actions and then use a machine learning algorithm to learn to discriminate between the two. Unfortunately, the collection and annotation of a sufficiently large dataset has a considerable cost. A major research direction in computer vision is to derive algorithm that can solve problems like ours in a completely unsupervised way, \ie, without the use of any labelled training data.

We solve this problem by adding knowledge to our system. Specifically, we use the basic knowledge of self-propelled motion, Newtonian motion and their relationship to reason about intentionality of an action. We derive a simple unsupervised computer vision algorithm for the recognition of intent based on these concepts. This demonstrates how simple, common concepts can be used to design systems that can perform complex, high-level tasks even when large amounts of labelled training data are not available.


\section{Related works}
\textbf{Visual recognition of intent in human.} 
The mechanism of visual recognition of intent has been the interest of congitive and social science since 1960s, although its underlying behavioral and neural mechanism is still an open question. The seminal work from Heider and Simmel \cite{H&S} shows that human subjects can assign personal attribute (like intentionality) to abstract geometric shape when the object moves in a human-like manner. \cite{sartori2011cues} shows that body movement plays an important role in human intent recognition. \cite{luo2005can} studied the capability of infants attributing goals to human and non-human agents when the agent moves in a self-propelled manner, supporting the hypothesis that the part of the recognition capability is rooted in a specialized reasoning system activated based on the kinematic feature of the object's action. \cite{chambon2017neural,chambon2011they} showed that intent recognition involves in an interplay of the kinematic information of the agent and prior expectation of the agent's movement. 

\textbf{Visual recognition of intention in computer vision.} Although significant progress has been made in some vision tasks like face/object recognition, there is very few studies focusing on visual recognition of intent of agent. \cite{wei2018and} proposed a hierarchical graph that jointly models attention and intention from a RGB-D video of an agent. But the study was focusing on the intention behind the eye gaze (the definition of attention in the study). \cite{vondrick2016predicting} proposed an algorithm to infer the motivation of the agent from an image with common knowledge factor graph extracted from text. \cite{ravichandar2017human} introduces an algorithm of estimate agent intention from the 3D skeleton of the upper body of the agent. In the study the intention is represented by latent state space defining the location of agent's arms, whose dynamic is defined by a neural network. This latent variable is then estimated by Expectation-Maximization (EM) algorithm. \cite{ullman2009help} developed an algorithm to infer a binary goal (help or hinder) of in a multi-agent setup with inverse planing in Markov Decision Process (MDP). Most of these works are based on a data-driven supervised model, which requires a large amount of labeled training data. 

Another area of research that is also related to the visual recognition of intent is human action/motion forecasting \cite{rudenko2019human}. Motion prediction aims at predicting actions from one or multiple agents in the future based on the observed actions in the past, where the intention recognition plays an important role (albeit very differently from the proposed study). \cite{fang2019intention} uses the 2D human pose to estimate pedestrians' intention of crossing and cyclists’ intention of turning and stopping. \cite{varytimidis2018action}, which also addresses the problem of pedestrian crossing/non-crossing recognition, shows that among different combinations of handcrafted/deep features with data-driven learning models, CNN deep feature and SVM shows the best performance on Joint Attention for Autonomous Driving (JAAD) dataset.

Although the aforementioned works shared the name of ``intent recognition'' with our study, the task is however very different. First, the purpose of this study is to recognize intentionality, i.e., recognizing whether an observed action is performed intentionally or not, rather than predicting the future human behavior based on a confined set of actions. In other words, our study is focusing on understanding the past, rather than predicting the future. Second, the previous works focusing on recognizing different intentions, with the assumption that all actions from an agent are intentional. However, this assumption might not hold for an arbitrary action (the action might be non-intentional), which can be tested by our algorithm. To our knowledge, there is no published computer vision system for the recognition of intentional/non-intentional action. 

\begin{figure*}
\includegraphics[width=1.0\textwidth]{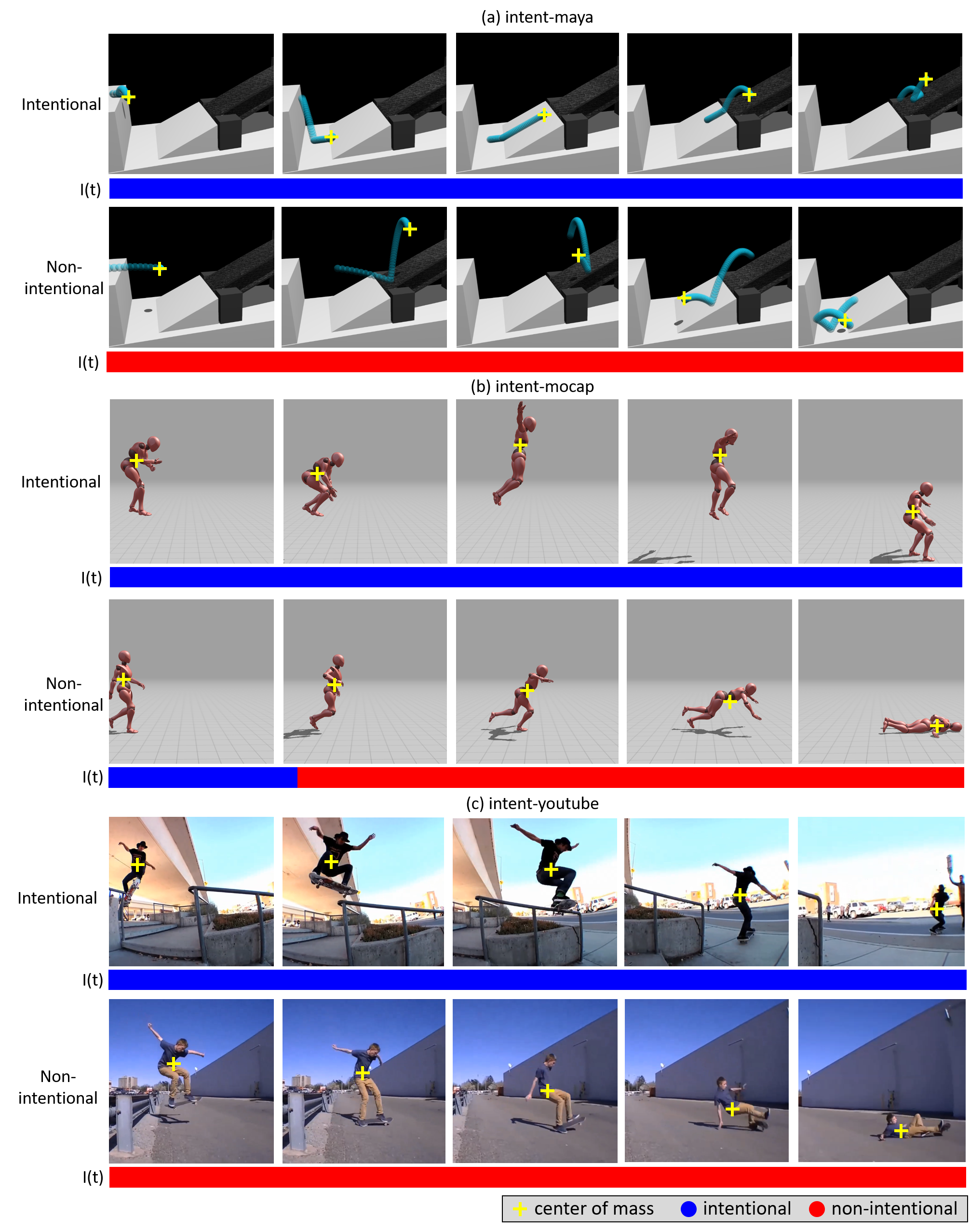}
\caption{Recognizing intentional versus non-intentional actions. The six samples are from the three datasets introduced in Section \ref{ss:datasets}. The colored horizontal bar underneath each image sequence denotes an intentionality function $I(t)$ of the action. The yellow crosshair illustrates the 2D project of the location of agent's center of mass. (a) intent-maya dataset. The intentional action showing a ball agent jumping down from a platform and climbing up a conveyor belt. In the non-intentional action, the ball moves according to the Newtonian physics. The transparent tail of the ball shows the location of the agent in the last second. (b) intent-mocap dataset. In the intentional action the agent jumps down from a (invisible) platform. In the non-intentional action the agent trips while walking. These snapshots of animation is directly extracted from www.mixamo.com. (c) intent-youtube dataset. In the intentional action, the agent successfully completed a board slide. In the non-intentional action, the agent falls at the end of an ollie.}
\label{fig:dataset-samples}       
\end{figure*}

\textbf{Common knowledge in computer vision.}
Incorporating common sense knowledge in computer vision system is also a largely unexplored territory in the community. \cite{aditya2015visual,del2013common} proposed rule based commonsense reasoning systems for visual scene understanding and action recognition, but with a focus on only hand related actions. \cite{zellers2018recognition} introduced the task of so called ``visual commonsense reasoning'' with corresponding dataset, where the machine is asked not only to answer question about the action and interaction between agent, but also the rationale behind such action. The rationale of an action is not directly observable in the given image, thus must be inferred through commonsense reasoning.

\section{Visual Recognition of Intent}
\label{section:method-modeling}

\subsection{Problem Formulation}
\label{section:problem-formulation}
Our goal is to design an unsupervised computer vision system that can classify observed actions of an agent or object as intentional or not. Given the trajectory of the agent's (object's) center of mass, we would like to parse the trajectory into segments that either exhibit intentional movement or unintentional movement. 

Let the 3D location of the agent (object) as a function of $t$ be denoted by $\mathbf{p}(t)=(x(t),y(t),z(t))^T$, with $y(t)$ indicating the vertical axis pointing up (\ie, up defines the positive quadrant). 

We now define the intentionality of the action of the agent as $I(t)\in\{1,-1\}$, with 1 indicating the action is intentional and $-1$ non-intentional; note $I(\cdot)$ is also a function of time, since some parts of the observed action may correspond to intentional actions (\eg, walking), while others to non-intentional (\eg, lose one's footing). 

Hence, our goal is to construct a model $f(\cdot)$ such that $I(t)=f(\mathbf{p}(t))$. Since we wish to do so without any training or the need for labeled data (\ie, an unsupervised approach), herein, we derive a model of $f(\cdot)$ which incorporates common knowledge about intentional and non-intentional behavior of an agent. 

Figure \ref{fig:dataset-samples} provides six examples of this task, ranging from animations of abstract geometric objects (intent-maya, Figure \ref{fig:dataset-samples}(a)), to animations of humanoid characters (intent-mocap, Figure \ref{fig:dataset-samples}(b)), then to real-world video of human actions (intent-youtube, Figure \ref{fig:dataset-samples}(c)). The colored horizontal bar in Figure \ref{fig:dataset-samples} denotes an $I(t)$ of an action. Our task is to construct a model that maps the 3D trajectory of the agent's center of mass (shown as yellow crosshairs in Figure \ref{fig:dataset-samples}) to the intentionality of the agent's action $I(t)$ (blue/red horizontal bar in Figure \ref{fig:dataset-samples}). 

\begin{figure*}
\includegraphics[width=1.0\textwidth]{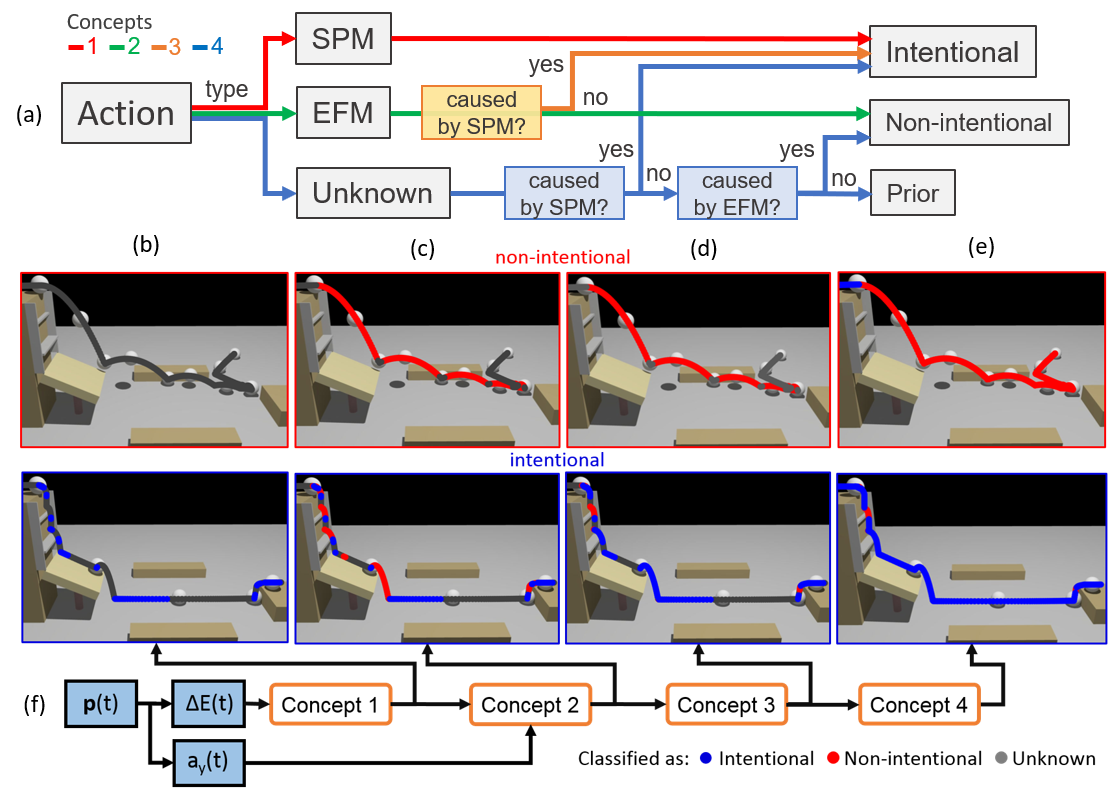}
\caption{Overview of the proposed algorithm. Here we illustrate the concepts we derive to model intentionality. (a) shows a logic diagram of the four concepts introduced in Section \ref{ss:concept-intro}, and their relationship with intentionality. (b-e) shows a pair of samples from our dataset described in Section \ref{sss:intent-maya-description}. The intentional example (in the blue box) shows a ball stepping down a ladder and jumping down an inclined platform to go to the isle at the far end of the scene. In the non-intentional example, the ball rolls and bounces according to Newtonian physics, with a trajectory that closely mimics that of the intentional action, yet the human eye is not trick by this and people clearly classify the first action as intentional and the second as non-intentional. (b) Result of our algorithm when  only Concept 1 is considered; (c) result with Concepts 1 and 2; (d) result with Concepts 1, 2 and 3; (e) results with all four concepts included in our algorithm; (f) model overview. The proposed algorithm first extract change in total mechanical energy $\Delta E(t)$ and the vertical acceleration $a_y(t)$ from the input trajectory of the agent, $\mathbf{p}(t)$.  Concept 1 recognizes intentional action from $\Delta E(t)$. Concept 2 takes $a_y(t)$ and the output of Concept 1 to form an understanding on non-intentional actions, which will be used in Concept 3 to update the decision. Finally,  Concept 4 handles all the unknown state that is previously unrecognizable (see derivation in the main text of the paper for details).}
\label{fig:2}       
\end{figure*}

\subsection{Common knowledge concepts}
\label{ss:concept-intro}
Imagine a human agent jumping over a hurdle, which is clearly an intentional action. When she prepares to jump, she converts the (non-observable) chemical energy stored in her body to the mechanical energy of her muscle. The muscle contracts and pushes her body upward in the air. While in the air, gravity is the main external force acting on her which forces her to fall back to the ground. If the initial muscle contraction is strong enough, she successfully jumps over the hurdle. 

If we examine the total mechanical energy of the system in the above example, which includes the scene and the agent, we see stable energy before the jump, a sharp increase at the time of the jump, and a stable trend after the jump (during free fall back down). For us human, the association between the perception of intentionality and the function of total mechanical energy is among many common knowledge concepts that we gradually learn in the early stage of the development of our brain \cite{luo2005can}. Our goal is to incorporate this knowledge into a computer vision system, thus avoiding the need to train a supervised machine learning algorithm to model intentionality from labelled data. 

This study models the following common knowledge concepts, 
\begin{itemize}
\item \textbf{Concept 1 (C1):} A standalone\footnote{Standalone means this concept only focuses on the movement at a specific time point rather than the relationship between actions.} self-propelled motion (SPM) is an intentional action, where self-propelled motion (SPM) is any movement that adds observable mechanical energy into the system.

\item \textbf{Concept 2 (C2):} A standalone external-force motion (EFM) is a non-intentional action, where the external-force motion (EFM) is any movement induced only by external forces (e.g., gravity).

\item \textbf{Concept 3 (C3):} An EFM caused by a SPM is part of an intentional action (\eg, falling down after an upward jump). 

\item \textbf{Concept 4 (C4):} An agent has inertia of intentionality (II), meaning the intentionality of an agent does not change unless C1-C3 applies. 
\end{itemize}

The four concepts and their relationship with the intentionality of an action can be visualized by Figure \ref{fig:2}(a) in the form of a logic diagram. 

Similar to Newton's Three Laws of Motion, any of these concepts alone does not fully define intentional/non-intentional actions across time. Only when combined, they form a common knowledge system that can be used to recognize intentionality for an agent across time. 


\subsection{Mathematical derivations}
\label{s-section:math-formulation}
Recall that we want to formulate the common knowledge as a functional mapping $f(\cdot)$, such that $I(t) = f(\mathbf{p}(t))$, where $I(t)$ and range of $\{ -1, 1\}$ at each time $t$. During the definition of each Concept 1 and 2, we will also use 0 to denote an ``unknown'' state, which is an intermediate state that will be categorized in Concept 4. 

\subsubsection{Concept 1}
\label{ss-section:C1}
Concept 1 (C1) states that a standalone SPM is an intentional actions since SPM adds total mechanical energy to the observable system. C1 derives from the common knowledge that human utilizes internally stored energy to execute movements that fulfill his/her intention, adding observable energy into the system. Thus the model of C1 can be derived as follows,
\begin{equation}
\label{eq:model-C1}
    I_{C1}(t) = f_{C1}(\mathbf{p}(t)) =\left\{\begin{array}{l}1\;\;\;\mathrm{if}\;\Delta E(t)>0\\
                                                       0\;\;\;\mathrm{otherwise,}\end{array}\right.
\end{equation}
where $\Delta E(t)={dE(t)}/{dt}$ is the change in the total observable mechanical energy $E(t)$ with respect to time, 
\begin{equation}
\label{eq:mechanical-energy}
E(t)=K(t)+V(t),
\end{equation}
with $K(t)$ the kinetic energy given by, 
\begin{equation} 
K(t)= \frac{1}{2}\left[\left(\frac{dx(t)}{dt}\right)^2+\left(\frac{dy(t)}{dt}\right)^2+\left(\frac{dz(t)}{dt}\right)^2\right],
\end{equation}
$V(t)$ the potential energy defined as,
\begin{equation}
V(t) =G\left(y(t)-y(t_0)\right),
\end{equation}
$G$ is the gravitational constant, and $y(t_0)$ is the initial y-axis location of the agent. In this formulation, we model agents as points with unit masses, neglecting the rotational kinetic energy or elastic potential energy.

$I_{C1}(t)$ will be equal to 1 at any instance in which the trajectory adds energy into the observable system and 0 to any other movement that does not specified in this concept.

\subsubsection{Concept 2}
\label{ss-section:C2}
Concept 2 (C2) states that a standalone EFM, a motion introduced by only external forces, is non-intentional. This is due to the fact that the exertion of the external force does not change depending on agent's desire or belief. For example, if an agent is falling, it is generally not the intention of the agent to be falling but, rather, the agent has no control over the effect of gravity, making this downward motion inevitable and, thus, non-intentional \cite{sep-action}. However, one should also notice that an agent may take advantage of the EFM, intentionally position themselves in the EFM to achieve their purpose. This special condition will be considered in concept 3. 

In practice, the number and types of external forces vary depending on the scene. But on earth, gravity is the primary external force we are bound by and, hence, this is what we are focusing on in the present work. 

There are two characteristics of gravity: 1. It is approximately equal regardless of the location of the agent, thus introducing a constant downward acceleration ($g$); 2. the effect of gravity on an agent (or object) does not increase the observable total mechanical energy of the system. The former will be modeled by $f_{C2g}(\mathbf{p}(t))$, while the latter is already modeled by Concept 1 and represented in $I_{C1}(t)$. 

With this knowledge, we can derive the model of C2, $f_{C2}$, as, 
\begin{equation}
\label{eq:model-C2}
\begin{aligned}
    I_{C2}(t) & = f_{C2}(\mathbf{p}(t), I_{C1}(t)) \\
              & = f_{C2g}(\mathbf{p}(t)) + I_{C1}(t),
\end{aligned}
\end{equation}
where $f_{C2g}(\mathbf{p}(t))$ is defined as, 
\begin{equation}
\label{eq:model-C2g}
     f_{C2g}(\mathbf{p}(t)) = \left
     \{
     \begin{array}{ll}
     -1 & \; \mathrm{if} \; I_{C1}(t)=0 \; \wedge  a_y(t)=c\cdot g, \\
     \; & \; \exists \; c > 0\\
     0 & \; \mathrm{otherwise,}
     \end{array}\right.
\end{equation}
$g$ is the negative acceleration due to gravity, $\wedge$ is the Boolean AND operation, and $a_y(t)$ is the vertical acceleration of the agent, which is defined as,
\begin{equation}
a_y(t) = \frac{d^2y(t)}{dt^2}.
\end{equation}
Note that we defined the $y$-axis to be pointing vertically upward, opposite to the direction of gravity. 

The condition for $-1$ (non-intentional) in the equation (\ref{eq:model-C2g}), $a_y(t)=c\cdot g, \; \exists \; c > 0$ represents a downward, constant acceleration. The other condition $I_{C1}(t) = 0$ showing the movement does not add anything to the total mechanical energy of the observable system. The two conditions are combined with an AND operator to ensure that both are simultaneously satisfied. 

One may wonder why $c>0$, rather than $c=1$, meaning the vertical acceleration of the agent is equal to the gravitational acceleration on earth. The reason is that by having $c>0$ we can model the motion due to gravity when the agent is on an inclined surface. The case that the object moved in a uniform speed ($c=0$) is assigned to the unknown state under this concept since the motion is not due to mere gravity.

Equation (\ref{eq:model-C2}) adds $f_{C2g}$ and $I_{C1}$ together, which gives 1 for intentional, -1 for non-intentional, and 0 for all the unknown movement that are not described by either C1 or C2. Those unknown movements will be handled by Concept 4. 

\subsubsection{Concept 3}

\begin{figure}[t]
\centering
  \includegraphics[width=0.45\textwidth]{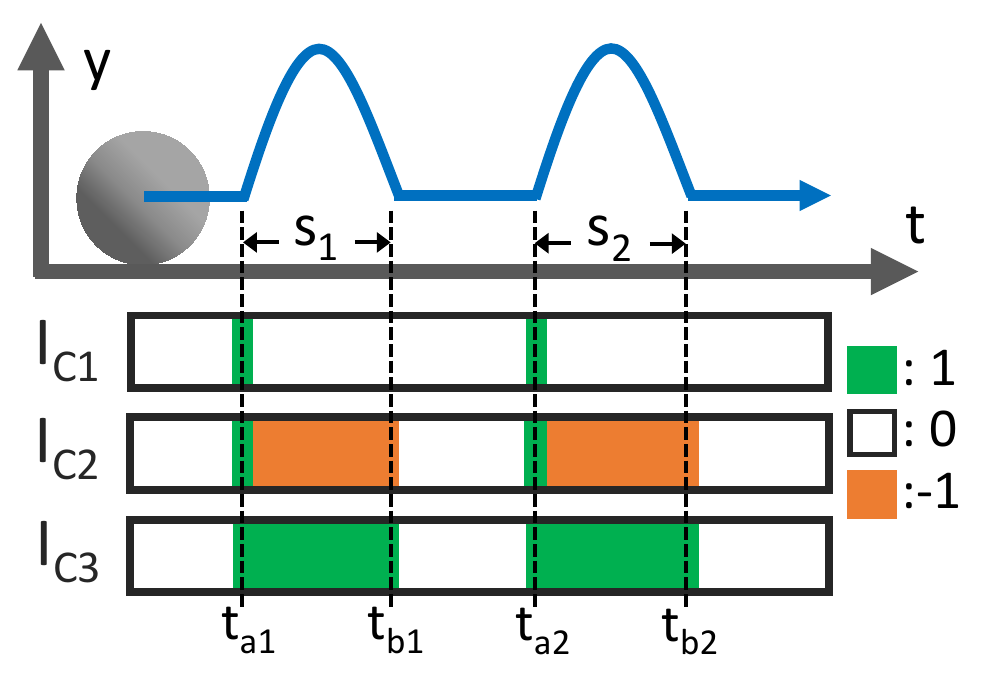}
\caption{An example where Concept 3 is necessary to achieve a correct classification of intentionallity. In this trajectory, an agent jumps twice. $I_{C1}$, $I_{C2}$ and $I_{C3}$ is the output of Concepts 1, 2 and 3, respectively. At time $t_{a1}$ and $t_{a2}$, the agent adds positive energy into the system to initialize the jumps. Thus, the movement at these two time points is detected as intentional as shown in $I_{C1}$. $s_1$ and $s_2$ are the two time intervals when the agent's movement is induced only by gravity, i.e., free fall. Hence, the action in these two intervals is detected as non-intentional by $I_{C2}$. However, since the free fall is part of the jump, the correct classification should be intentional. By taking into account causal relationship between action, Concept 3 can correctly classify these two movement as intentional, as shown in $I_{C3}$.}
\label{fig:example-C3}       
\end{figure}

Concept 3 (C3), as foreshadowed in Section \ref{ss-section:C2}, describes the condition that an EFM might not be non-intentional when the agent actively moves herself to the status of EFM. For example, when the human agent was jumping over the hurdle mentioned earlier in this section, she was subjected to gravity forces after she pushes herself in the air. Although the free fall motion is induced by mere gravity, the motion is nevertheless the result of her initial jump -- an intentional action that adds total mechanical energy into the system. C3 is modeling exactly this condition, when a EFM is casused by a SFM, this EFM should be classified as intentional movement. 

However, modeling the causal relationship between actions is a challenging problem by itself. In this study, we simplify the causality to an immediate temporal relationship, i.e., the causal action is immediately before the consequential action, which is a surrogate we found works well. Temporal precedence is one of the criteria that is necessary for constructing causality. The reason we only focus on short-term causality is that the long-term causal relationship between actions can be decomposed to a chain of short-term causal relationships between actions. 

To model this knowledge, let us first define the set of time intervals of all EFMs as $S_{\mathrm{EFM}} = \{s_1,s_2,...,s_i,...\}$ whose elements $s_i$ are the time interval of the $i^{th}$ EFM, as shown in Fig \ref{fig:example-C3}. The main idea of the algorithm is, for each EFM, identify if it is caused by a SFM. If so, EFM will be recognized as an intentional action. More formally, the model $f_{C3}$ is formulated as shown in Algorithm \ref{alg:C3}.

\begin{algorithm}
 \caption{Algorithm for C3}
 \label{alg:C3}
\SetAlgoLined
 Input: $I_{C1}, S_{\mathrm{EFM}}$ \;
 Output: $I_{C3}$ \;
 Initialize $I_{C3} \gets I_{C1}$ \;
 \For{ $s_i \in S_{\mathrm{EFM}}$}{
    $t_{ai} \gets \mathrm{inf}(s_i)$ \;
    \If{$I_{C1}(t_{ai}-1)==1$}{
    $I_{C3}(t)\gets1$ for $\forall t \in s_i$\;
    }
   }
\end{algorithm}

In Algorithm \ref{alg:C3}, $t_{ai} \gets \mathrm{inf}(s_i)$ extracts the starting time point for the $i^{th}$ EFM. The operation of $I_{C1}(t_{ai}-1)==1$ examine if the movement immediate preceding EFM is a SFM. In such case, SFM is treated as the cause of EFM, which means EFM is also intentional. The assignment of intentionality is implemented as $I_{C3}(t)\gets 1$, for $\forall t \in s_i$ in the algorithm. 

Figure \ref{fig:example-C3} illustrates an example of the case where C3 is needed for correct recognition of intentionality. There are two EFMs in the figure, whose time intervals are denoted by $s_1$ and $s_2$. There are two instances of SPMs, which can both be abstracted as force impulse generated by the agent that initializes a ``jump''. Since the instantaneous nature of the impulse, our C1 model can only detect the SPM at two time points, $t_{a1}$ and $t_{b1}$ shown in the $I_{C1}$ row in the figure. The two EFMs, which are free fall in this case, is a direct and expected result from the initial SPM, thus should be treated as intentional.

\subsection{Concept 4}

Concept 4 (C4) is introduced to handle intentional movements that are not modeled by C1, C2 and C3. Using the concept of inertia from physics, which describes a resistance of the object to change its velocity, we describe C4 as an {\em intentionality inertia}, a property of the agent that resists changes in its intentionality status -- the intentionality of an agent does not change unless the one or more of concepts 1 through 3 occur. 

The rationale behind this concept can also be understood from the causal relationship of the actions. When a movement causes another movement, the intentionality carries over. However, if an event happens that breaks the causal relationship, in our case those event defined by the C1 to C3, the intentionality will change accordingly. Let us imagine a case in which a human agent falls from a cliff, hits the ground and lies on the ground since then. The unfortunate fall is a non-intentional movement, according to C2. The movement (or lack of movement) of lying on the ground is also non-intentional. It is not the agent's intention to fall at the first place, so it is also not the intention of the agent to be lying on the ground since lying on the ground is an effect of the falling and hitting the ground. Thus, although ``lying on the ground'' is not one of the actions defined in C1 to C3 (does not add total mechanical energy; does not have a constant downward acceleration), it is still non-intentional due to its relationship to its cause action. If the agent standing up after lying on the floor, the ``standing up'' will be recognized as intentional according to C1. 

To model this concept, we first define a set of time interval of all the ``unknown'' actions, $U_{\mathrm{null}} = \{u_1,u_2,...,\\u_i,...\}$ whose element $u_i$ is the time interval of the $i$-th unknown movement - the ones that does not belong to C1 to C3. For each of the unknown movement, we check the intentionality of the previous action, and assign the previous intentionality state to the current unknown action. More formally, the concept is formulated in algorithm \ref{alg:C4}.

\begin{algorithm}[h!]
 \caption{Algorithm for C4}
 \label{alg:C4}
\SetAlgoLined
 Input: $I_{C3}, U_{\mathrm{null}}$ \;
 Output: $I_{C4}$ \;
 Initialize $I_{C4} \gets I_{C3}$ \;
 \For{ $u_i \in U_{\mathrm{null}}$}{
    $t_{ai} \gets \mathrm{inf}(u_i)$ \;
    $I_{\mathrm{cause}} = I_{C3}(t_{ai}-1)$ \;
    $I_{C4}(t)\gets I_{\mathrm{cause}}$ for $\forall t \in u_i$\;
   }
\end{algorithm}

One may wonder what if the unknown movement happens at the beginning of the video where there is no C1-C3 motion defined as cause. In those cases, prior knowledge about the nature of the agent is needed, i.e., the assumption about the default intentionality of the agent. For a human agent, one might want to assume the default state is intentional, since the action from a normal, conscious adult is generally intentional by default (otherwise there is no reason for that person to move). In the case that no prior knowledge is available, the algorithm will output the unknown states. 

Now that we derived all the implementation of concepts 1-4, the final $I(t)$ is directly equal to $I_{C4}(t)$. Note that although $I(t)=I_{C4}(t)$, $I(t)$ is also a combination of all four concepts, since $I_{C4}(t)$ is a function of $I_{C3}(t)$ which itself is a function of both $I_{C1}(t)$ and $I_{C2}(t)$ (shown in Algorithm \ref{alg:C3} and Algorithm \ref{alg:C4}). 

\subsection{Implementation Detail}
To apply our algorithm on the trajectories with discrete time (frames), we use 1st order finite difference to approximate the derivative. We applied a 30-frame median filter on the estimated total mechanical energy from equation (\ref{eq:mechanical-energy}) to remove outliers. The condition that compared to zero in equation (\ref{eq:model-C1},\ref{eq:model-C2g}) will replace zero with a positive threshold close to zero to account for possible numerical error in the estimation. 

\section{Assumptions of the system}
In this section, we will give a thorough analysis on the assumptions of the proposed algorithm. We will delineate the assumptions made on the Computational Level (Section \ref{ss:concept-intro}) and the Algorithmic Level (Section \ref{s-section:math-formulation})\footnote{Here we are using the Computational, Algorithmic, and Implementational level from David Marr \cite{vision1982marr}. The implementational level is not discussed since our work does not contribute to that specific level}. 

We argue that the proposed common knowledge concepts are relatively general on the computational level, meaning that the logic described by the four concepts are generally sufficient to determine intentionality across layers of abstraction (the same logic can be applied to Heider-Simmel-like animations as well as real word videos). However, the algorithmic implementation we used is based on a set of assumptions which limits its generality. To explain this, let us consider an example of a cue ball hitting a pool ball. In this example, when the interested agent is the pool ball, the movement is due to an external-force movement, induced by the impulse from the cue ball. Because this EFM is not a result of a self-propelled motion, the algorithm, on the conceptual level, should correctly classify the movement as non-intentional. But, on the algorithmic level, our specific algorithmic implementation is telling a different story. Because the ball gains mechanical energy (through the impulse from the cue ball), the algorithm determines that the pool ball is performing self-propelled motion, and thus annotate the movement as intentional. 

The reason for the mismatch, is because of the following assumptions our algorithm operates on: 
\begin{enumerate}
    \item There is only one agent involved in the action. 
    \item The total mechanical energy of the agent can be calculated from its kinematics of the center of mass. 
    \item The external force is gravity and its decomposition. 
    \item The causal relationship between SPM and EFM can be described with immediate temporal relationship. 
\end{enumerate}

The four assumptions listed here might lead to the impression that the proposed system is very constrained. This might be true compared to the generic intention recognition, which is extremely complex and even humans fail to perform this task in cases. However, under the condition of action from a single agent in a static environment on earth, the set of assumptions are generally applicable, or could be approximated well enough for the algorithm to perform well, which we will show in the experimental results in Section 5. The clear presentation of assumptions, we argue, should be considered as an advantage rather than a weakness, since it allows practitioners to be aware of the condition where our algorithm is not applicable, and provides researchers clear future directions of improvement. 

\section{Experiments}
\label{Experiments}
To our knowlege, there is no existed dataset on intentional\//non-intentional actions. Thus, we created three datasets for our experiment: intent-maya, intent-mocap and intent-youtube. Intent-maya dataset contains abstract minimalistic 3D animation for intentional/non-intentional actions, providing 3D ground truth trajectory for sphere-like agent. Intent-mocap dataset contains motion capture data collected from human agents, providing accurate 3D location of human body but left the center of mass trajectory subject to estimation. Intent-youtube dataset provides in-the-wild RGB video samples where 3D location of human body and center of mass are both estimated. Although we provide manual labels of intentionality on all the three datasets, those labels are not used as part of our proposed algorithm, since our algorithm is not data-driven thus has no need for manual labels. The label is only used to evaluate the performance of the proposed algorithm and train the supervised baselines for comparison. Testing on these three datasets shows the capability of our algorithm to recognize intentionality in both abstract, idealistic dataset and realistic, noisy dataset, showing the general applicability of the proposed concepts. 

\subsection{Datasets}
\label{ss:datasets}
\subsubsection{Intent-maya dataset}
\label{sss:intent-maya-description}
Intent-maya datasets contains 60 3D animations of agents acting intentionally or non-intentionally, half for each class. The animations are designed similar to the stimuli in the classical Heider and Simmel experiment \cite{H&S}, in which they showed that human attribute intentionality even to abstract geometric objects. In our videos, one or multiple balls move in a manually designed 3D scene. The movement is human-like in intentional videos and Newtonian in the non-intentional videos. We use Autodesk Maya 2015 to generate the videos. Keyframe animation is used for the intentional videos and Bullet Physical Engine is used for the non-intentional videos. Each video has 480 frames at 60 fps. 

To ensure the videos can be perceived as intentional or nonintentional, we asked 30 Amazon Mechanical Turkers to evaluate each animation, judging if the action is intentional or not. All the videos in the dataset has at least 90\% agreement among Turkers indicating a consistent intentionality perception across human subjects. 

Since all animation are manually coded, we can extract the ground truth 3D trajectory of the center of mass of the agent directly from Maya animation.

\subsubsection{Intent-mocap dataset}
The 3D manually designed animation we created in Maya provided abstract but yet compelling intentional/ non-intentional perception on balls. However, one may view the animation in intent-maya dataset as too abstract and simple to be generalized to practical condition. Intent-mocap dataset is created to mediate this concern. Motion capture data provides us actions performed by human agent with the advantage of direct measurement on 3D location of body markers, yielding accurate 3D trajectories of the joints of the agent. 

We collect mocap sequences from Adobe Mixamo dataset\footnote{https://www.mixamo.com/}, which provide a wide range of intentional and non-intentional mocap sequences that are cleaned by keyframe animators. We manually select the intentional and non-intentional sample based on the action description provided in the datasets. The description for intentional actions includes jumping, walking, running, climbing, etc. The description of non-intentional actions includes, falling, tripping, slipping, etc. With these description, we collected total 208 samples, half for each class. A sequence of 21-joint skeleton is extracted from the mocap samples using MATLAB. The range of length the sequence varies from 32 to 1219 frames. The sampling rate of the sequence is 60 Hz.

We directly use the 3D human joints location provided the mocap data. 

\subsubsection{Intent-youtube dataset}
The mocap dataset provides precise human action sequence. However, the nature of the mocap generally requires the actors to perform pre-scripted actions. Thus even if we collect non-intentional samples, one could argue that the actor is to ``pretend'' to be non-intentional\footnote{However, one should also notice that acting to be non-intentional does not mean the action and kinematics of the agent lacks the characteristic of the genuine non-intentional movement}. We introduce intent-youtube dataset to address this concern. 

The youtube datasets contains 1000 in-the-wild human action videos, among which 500 are intentional actions and 500 are non-intentional actions. The videos are collected by keyword searching in YouTube. For intentional video, the keywords are derived from ``action'' and ``activity'' in WordNet \cite{miller1998wordnet} and ConceptNet \cite{speer2017conceptnet}. Non-intentional keywords consist two part: adjective and action (e.g., ``accidental drop''). Besides the keywords extracted from WordNet and ConceptNet, we also used keywords that empirically effective, like ``fail'' for non-intentional actions. Only the videos with above 100 views are used in our dataset. Camera shot detection is applied to each video to ensure each video clip only contains one camera. The video clips with significant camera motion are also removed from the dataset. All video samples have at least one full body agent. The final videos varied between 51 and 299 frames in length. 

To verify that these video properly exhibit either an intentional or unintentional action, each video was classified into the intent or non-intent categories by a Amazon Mechanical Turker and then verified by an experienced annotator. All the videos with inconsistent judgment from the annotators are removed from the dataset. 

We extract 3D human pose by applying 3D human pose estimation algorithm proposed in \cite{martinez2017simple} on the 2D human pose extracted by OpenPose \cite{cao2018openpose}. Given a estimated 3D human pose, we solve a perspective n-point (PnP) problem with non-linear least square (with steepest decent algorithm) to estimate the 3D translation of the agent. 


\subsection{Estimating center of mass for human agent}
\begin{figure}[h]
\centering
  \includegraphics[width=0.48\textwidth]{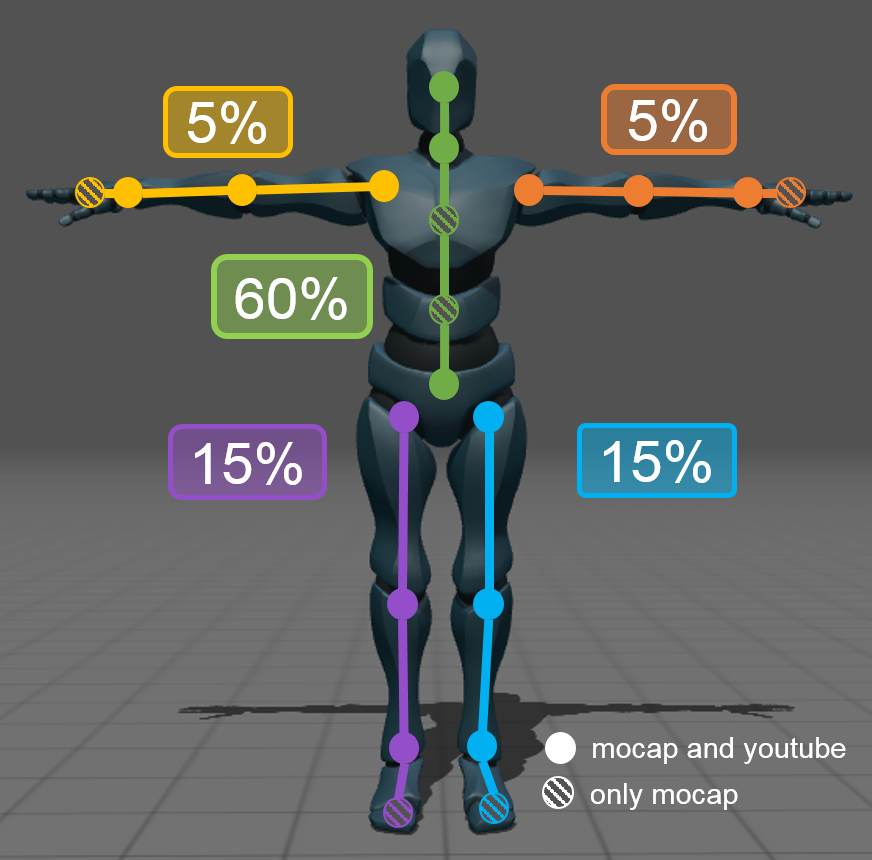}
\caption{Illustration of weight distribution used in calculating center of mass in the mocap and youtube datasets. The joints with solid color are used in both mocap and youtube skeleton template. The joints with diagonal pattern are only used in the mocap skeleton template.}
\label{fig:weight-distribution}       
\end{figure}

To estimate the center of mass of the human agent, we first assign each joint to either legs (from hip to the toe), torso (lower back, spine, lower spine and head) and arms (shoulder, elbow, wrist and hand). Then the center of mass of each human body component was computed by averaging all the points assigned to the body part. The center of mass of the agent is then calculated by weighted averaging the body part center, with the weight defined by the standard human weight distribution \cite{HumanBodyDynamics}, see Figure \ref{fig:weight-distribution}.

\subsection{Recognition of intent in videos}
\label{ss:recognition-method}
The algorithm introduced in Section \ref{section:method-modeling} recognizes intentionality in each segment for a single agent, which has to be aggregated for the final intentionality label for the entire video. 

For the samples in intent-maya datasets, we know that each video either contains purely intentional or purely non-intentional actions. Thus if the number of detected intentional segment is greater than the nonintentional segments, the video is intentional, and \textit{vice versa}. More formally, the final decision for the video, $I_v$ is defined as follows,
\begin{equation}
    I_v = \left\{\begin{array}{ll}\mathrm{intentional} \; & \mathrm{if}\;\sum_{j = 1}^N \sum_{t=1}^{T_j} I_{C4,j}(t)>0\\
                                 \mathrm{nonintentional} \; & \mathrm{otherwise}
                 \end{array}\right.
\end{equation}
where $I_{C4,j}(t)$ denotes the result of C4 for the $j$-th agents in the video. $\sum_{t=1}^{T_j} I_{C4,j}(t)$ denotes the difference between the number of intentional segments versus non-intentional segments for $j$-th agent, which is calculated by summation since intentional action is labeled as 1 and non-intentional as -1. 

Unlike the sphere-like agent in the intent-maya dataset, the non-intentional action of human agents is usually happen in the middle of intentional actions (e.g., a human slips in the middle of a walk, with walking as intentional but slipping as non-intentional). Thus we recognize the action of the agent as non-intentional if the number non-intentional segments is above a threshold (we set threshold equal to 40 frames in our experiment). Otherwise the action of the agent in the video is intentional.

\subsection{Comparison between our algorithm and baseline methods}

We first compare our algorithm against 4 baseline methods: Linear Discriminate Analysis (LDA), Nearest Neighbour (NN), Kernel Subclass Discriminant Analysis (KSDA) \cite{you2011kernel}, a deep residual network (ResNet) \cite{he2016deep}, a recurrent neural network with Long Short Term Memory modules (LSTM), and a recurrent neural network with attention mechanism (LSTM+attention). For the latter two baselines (LSTM and LSTM+attention) we also test their performance with RGB video as input rather than 3D trajectories with an additional baseline from R(2+1)D \cite{tran2018closer}. These collection of baselines represents a wide spectrum of methods ranging from simple linear method to modern deep learning based method. The performance on these methods will show the level of difficulties of the problem of intent recognition.

\subsubsection{Implementation detail for LDA, NN and KSDA}
\label{sss:detail-baseline-classic}
For LDA, NN and KSDA, we first applied a 30-frame sliding window with a step size of 15 to the trajectory of each agent. We then used Discrete Cosine Transformation (DCT) to map each x, y, z component of the trajectory segment to a 10 dimensional space of DCT coefficients, which defines a 30 dimensional feature space. Samples extracted from different agents are pooled together to form the training set. During testing, the classification is first conducted on the segment level and then the same thresholding method is used as in Section \ref{ss:recognition-method}. 10-fold cross-validation is used to partition the datasets to training and testing. 

\subsubsection{Implementation Detail for ResNet}
\label{sss:detail-baseline-resnet}
For ResNet, instead of handcrafting the feature space, we directly input 3D trajectory segment to the network and have the network learn the feature representation for classification. Same sliding window and cross-validation method is used as in the LDA, NN and KSDA. We used ResNet-18 with modification on the first convolutional layer and maxpooling layer to accomodate the input dimensionality of the 3D trajectory segment. The kernel size of the first convolutional layer is $7\times3$ with padding $=2$. For the fisrt maxpooling layer, the kernal size $=3$, stride $=2\times1$ and padding $=1$. We use Adam optimizer with learning rate $=.001$, $\beta_1=.9$ and $\beta_2=.999$. We trained the network in 100 epochs with batch size $=128$. Similar to the testing procedure in the Section \ref{sss:detail-baseline-classic}, for a given testing trajectory, the network is applied on the segments of the samples, giving binary classification result for each segment. The final decision for the video is given by the majority vote of all the segment results of the testing trajectory. 

\subsubsection{Implementation Detail for LSTM and LSTM+attention}
For LSTM \cite{hochreiter1997long}, we input the entire 3D trajectory of the agent to the network instead of the 30-frame segment as in previous baselines. This allows the LSTM baseline to learn to recognize video-wise intentionality from an entire trajectory, rather than using the simple hand-crafted rules for aggregating segment-wise inference as described in Section \ref{ss:recognition-method}. We use 10 dimensional hidden state and cell state, initialized with zero vectors at the beginning of each sequence. At the last frame, the hidden state is fed to a 10-by-2 fully connected network with softmax. The network is optimized using Stochastic Gradient Descent (SGD) with learning rate $=.001$ and momentum $=.9$ to minimize the cross-entropy loss. 

For LSTM+attention, we used the attention mechanism proposed in \cite{bahdanau2014neural}, which models temporal attention as a bi-directional LSTM with 10-dimensional hidden and cell states. We use the same LSTM model described above to model the trajectory dynamics, jointly optimized with attention module using cross-entropy loss. The network is also optimized using SGD with learning rate $=.001$ and momentum $=.9$. 

\subsubsection{Implementation Detail for video based classification}
We also provide three additional baselines, LSTM+ResNet, LSTM+ResNet+attention and R(2+1)D \cite{tran2018closer} with images sequences as input rather than 3D trajectories of agents. These baselines provide insight on the effectiveness of 3D trajectory as input feature. The image sequence is extracted at every 10 frames to reduce the total length. For each frame in this sequence, the RGB image within the bounding box of an agent is extracted, then resized to 224$\times$224. For both LSTM+ResNet and LSTM+ResNet+attention baselines, a 512 dimensional feature is extracted after the average pooling layer, which will be used as input feature to the LSTM module. The hidden and cell states of the LSTM used in this experiment are both 512-d to accommodate the increase in dimensionality of the input features. An 512-by-2 fully-connected network is used to recognize the action of given agent is intentional or not. Both ResNet18 and LSTM (with attention module) are jointly trained using SGD with learning rate $=.001$ and momentum $=.9$. R(2+1)D is trained using Adam optimizer with learning rate $=.001$, $\beta_1=.9$ and $\beta_2=.999$.

\subsubsection{Quantitative Result}

\begin{table*}[h!]
\centering
\caption{Quantitative result comparison between the ours and baseline algorithms, measured by mean classification accuracy and standard error of the mean (in parenthesis). 3D COM: 3D trajectory of the agent's center of mass}
\label{tab:baseline-compare}       
\begin{tabular}{l l l l l}
\hline\noalign{\smallskip}
Methods & input & maya & mocap & youtube\\
\noalign{\smallskip}\hline\noalign{\smallskip}
LDA             & 3D COM   & 0.533 (0.060)  & 0.755 (0.014)  & 0.653 (0.017)\\
NN              & 3D COM    & 0.683 (0.052)  & 0.805 (0.023)  & 0.654 (0.014)\\
KSDA            & 3D COM    & 0.633 (0.048)  & 0.795 (0.022)  & 0.577 (0.013)\\
ResNet          & 3D COM    & 0.783 (0.058)  & 0.760 (0.025)  & 0.580 (0.019)\\
LSTM            & 3D COM    & 0.581 (0.232)  & 0.835 (0.082)  & 0.615 (0.057)\\
LSTM+attention  & 3D COM    & 0.504 (0.209)  & 0.671 (0.155)  & 0.505 (0.059)\\
\hline
LSTM+ResNet             & RGB video & 0.550 (0.200)  & -              & 0.770 (0.036)\\
LSTM+ResNet+attention   & RGB video & 0.517 (0.089)  & -              & 0.704 (0.079)\\
R(2+1)D                 & RGB video & 0.500 (0.091)  & -              & 0.609 (0.017)\\
\hline
\textbf{Ours}   & 3D    & \textbf{0.950}  & \textbf{0.827} & \textbf{0.785}\\
\noalign{\smallskip}\hline
\end{tabular}
\end{table*}

Table \ref{tab:baseline-compare} shows the mean classification accuracy and its standard error for the four baseline methods in maya, mocap and youtube datasets. We use leave-one-pair-out cross-validation for the baseline experiments in maya dataset, and 10-fold cross-validation for mocap and youtube dataset. We did not calculate the mean accuracy for our method since our method does not need training thus the entire dataset is used as testing set without cross-validation. As shown in the table, the accuracy of our proposed algorithm is significantly higher than the accuracy of the baseline methods in intent-maya dataset. In intent-mocap and intent-youtube dataset our algorithm produces comparable results to the most accurate baseline methods. It is worth noting that our algorithm achieves these result without any supervision or training, comparing to all the baselines which are learning based methods. 

As demonstrated by the above experiments, the algorithm described in this paper is general and can be applied to any video of an action. To prove this further, we decided to apply our approach to a new dataset that appeared long after we had submitted the first version of this paper.\footnote{This experiment was added during the revision phase of this paper.} Thus, this serves as an independent test on a data we had no access to during the design of our algorithm. The database in question is the Oops! database \cite{epstein2020oops}, which shows a number of unintentional actions collected from YouTube. We thus used our derived algorithm to identify these unintentional actions in the dataset. In this challenging task, our algorithms achieved 66.51\% accuracy.

One may wonder why our result is significantly better than all the 3D baselines in the youtube dataset but only comparable to the best baseline in the mocap dataset, since they are both essentially the same type of data (3D human pose sequence). One possible explanation is related to the highly noisy samples. In intent-youtube dataset, the 3D trajectory of an agent is estimated from the 2D video rather than directly measured by sensors as in mocap dataset. This estimation process introduces significantly higher noise to the youtube dataset due to the limitation of the off-the-shelf 3D human pose estimation algorithm. When a powerful non-linear data-driven learning algorithm (like ResNet, KSDA and LSTM) is used to learn the underlying pattern in this dataset, it is more likely that the algorithm will overfit to the noise, ending up with higher testing error \cite{friedman2001elements}. Our algorithm, on the other hand, directly examines the kinematics feature without training, thus avoiding this possible issue. 

\begin{table}[h!]
\centering
\caption{Comparing leave-one-pair-out (LOPO) cross-validation versus 10-fold cross-validation on intent-maya dataset using LSTM and LSTM+attention.}
\label{tab:cv-compare}       
\begin{tabular}{l l l}
\hline\noalign{\smallskip}
cross-validation & LOPO & 10-fold\\
\noalign{\smallskip}\hline\noalign{\smallskip}
LSTM (3D)            & 0.581 (0.232) & 0.482 (0.103)\\
LSTM+attention (3D)  & 0.504 (0.209) & 0.365 (0.142)\\
\noalign{\smallskip}\hline
\end{tabular}
\end{table}

Table \ref{tab:cv-compare} also shows disadvantages of supervised methods in a biased dataset, which is almost always the case. This is particularly clear when examine the result of LSTM (3D) and LSTM+attention (3D) in intent-maya dataset using 10-fold cross-validation where the classification accuracy is even below the chance level of 50\%. The reason for the low performance on the maya dataset is its careful design. The samples in the maya dataset are designed in intentional-nonintentional pairs. Within a pair, the background, objects, and illumination agents are all identical. The only difference is the kinematics of the agent, which is also designed to be as similar as possible while preserving the significant difference in the perception of intentionality. When we randomly partition the dataset for 10-fold cross-validation, some intentional (non-intentional) samples in the validation set might have a their non-intentional (intentional) counterpart in the training set. Due to the data-driven nature of the supervised models, the similarity between the training and testing samples tends to bias the network towards the wrong decision. This is particularly true for the models like LSTM and LSTM+attention, which are given a higher flexibility to learn not only the features, but also the rules for video-wise recognition of intent. Our algorithm, on the other hand, does not have this disadvantage due to its common knowledge based inference. 

\subsection{Qualitative Result}

\begin{figure*}
\includegraphics[width=1.0\textwidth]{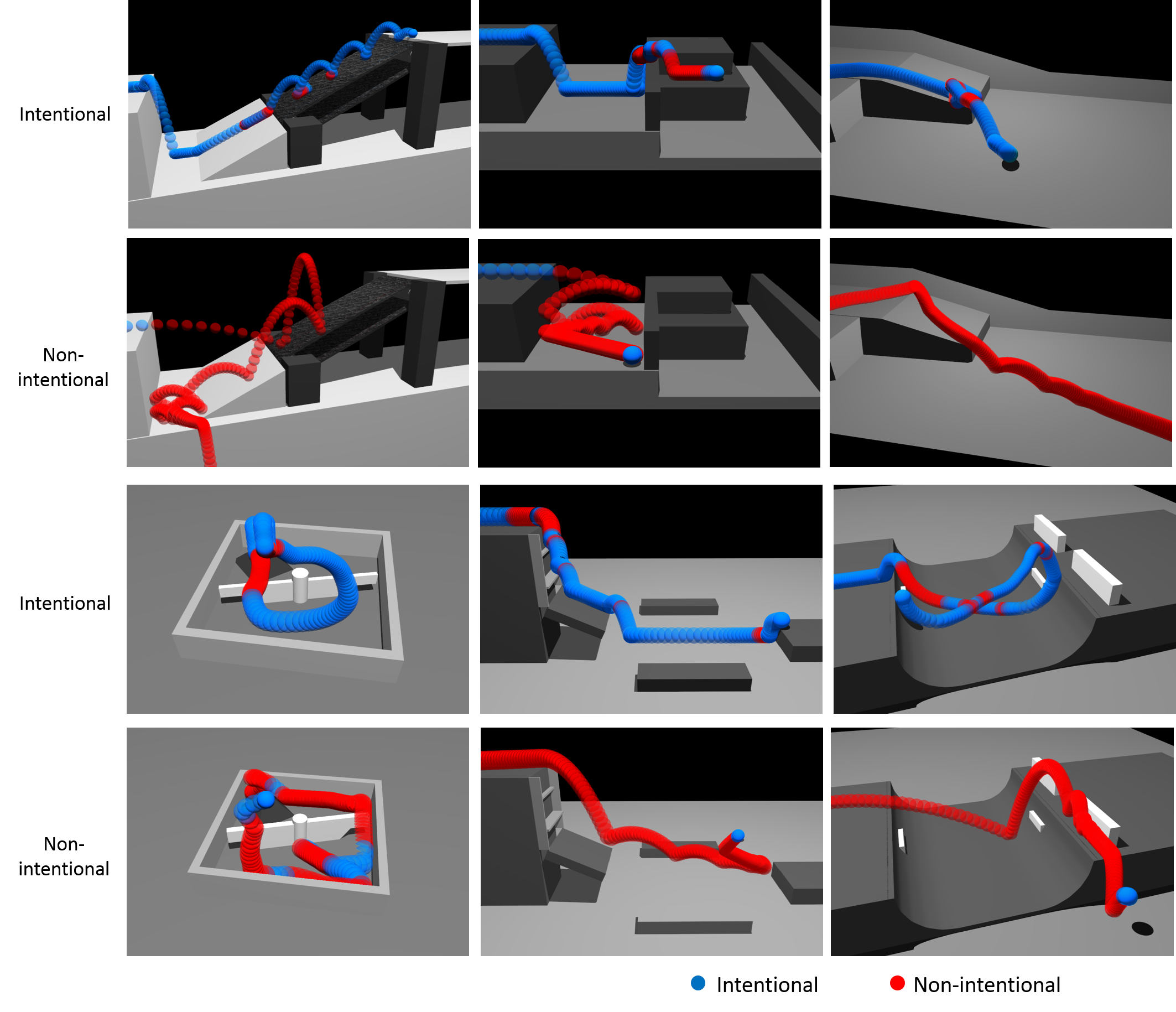}
\caption{Qualitative result of our algorithm testing on intent-maya dataset. The full model with all concepts is used. Blue (red) indicates that our algorithm recognizes the movement of the agent as intentional (non-intentional) at that specific time. The ground truth annotation is shown on the left of the figure.}
\label{fig:maya-result}       
\end{figure*}

\begin{figure*}
\includegraphics[width=1.0\textwidth]{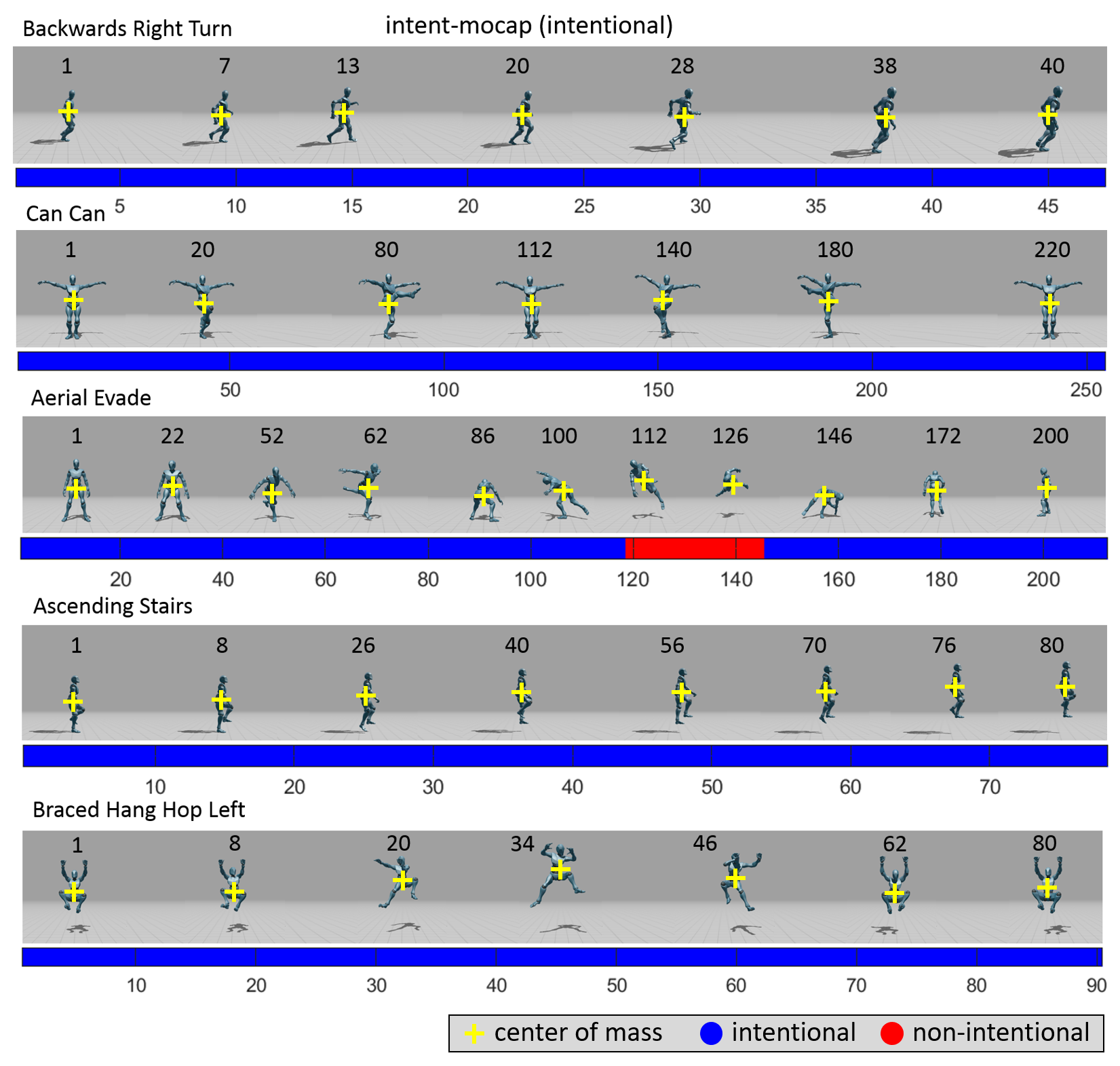}
\caption{Qualitative result of our algorithm testing on intent-mocap datasets. All samples shown here contains intentional actions. The full model with all concepts is used. The colorbar indicates the intentionality judgement by our algorithm at each frame, blue for intentional and red for non-intentional. The number above the agent is the corresponding frame index in the sequence. The action name is shown on the top-left corner of each sequence which corresponds to the animation name in mixamo dataset. We applied median filter on $I_{C4}$ with windows size 30 to increase smoothness or the result.}
\label{fig:mocap-intent}       
\end{figure*}

\begin{figure*}
\includegraphics[width=1.0\textwidth]{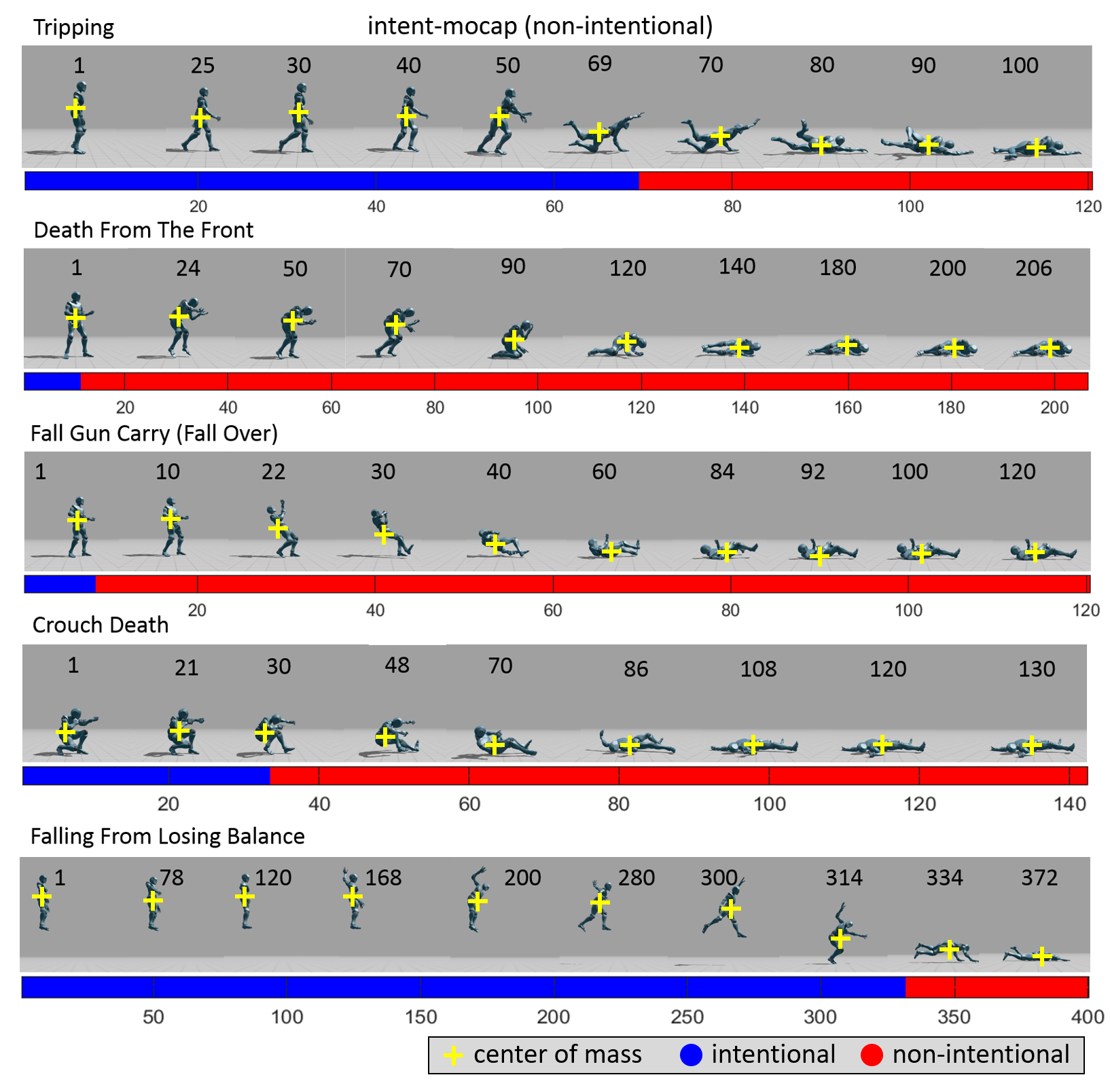}
\caption{Qualitative result of our algorithm testing on intent-mocap datasets. All samples shown here contains non-intentional actions. The full model with all concepts is used. The same method used in Figure \ref{fig:mocap-intent} is applied to generated this images. }
\label{fig:mocap-nonintent}       
\end{figure*}

\begin{figure*}
\includegraphics[width=1.0\textwidth]{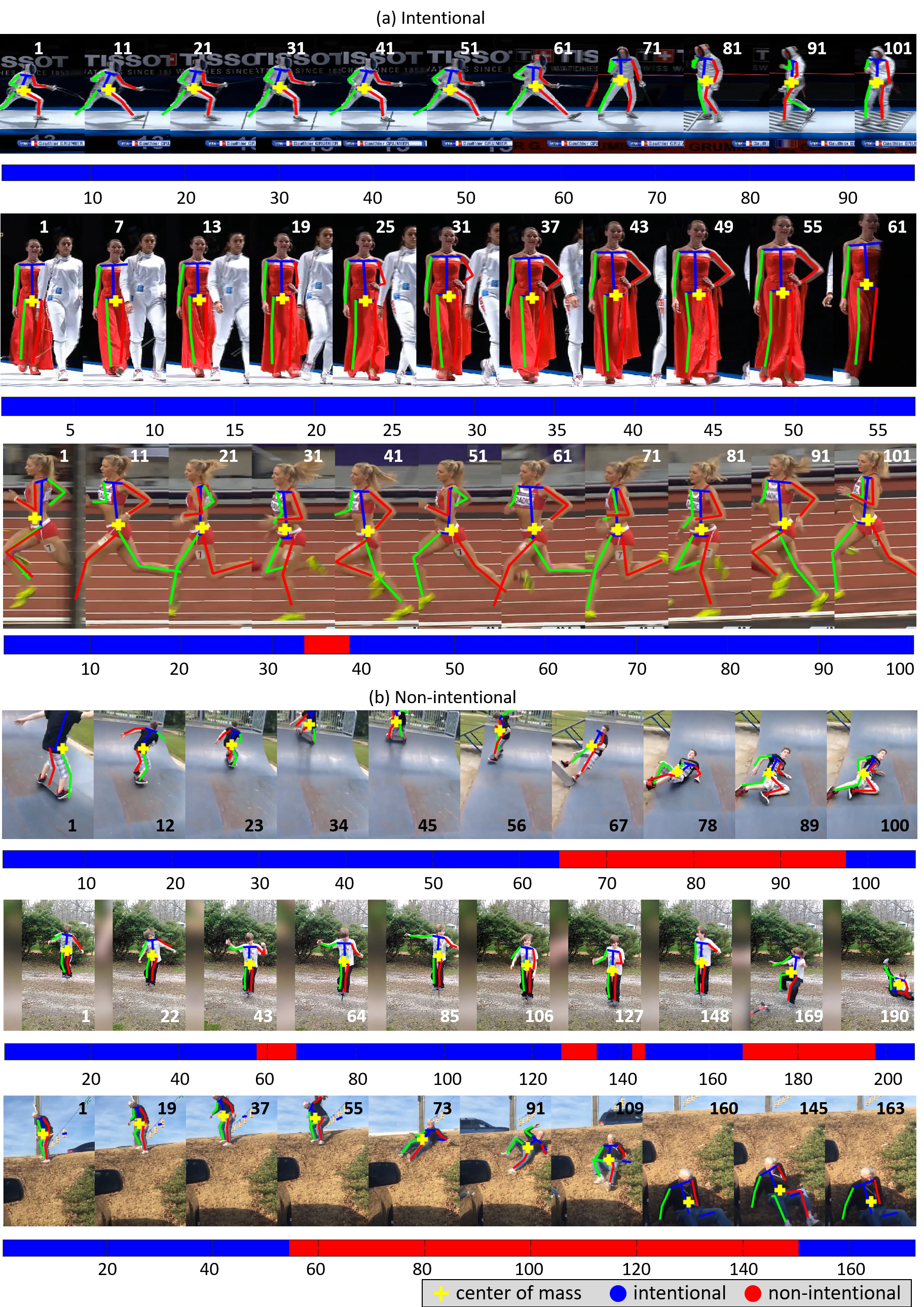}
\caption{Qualitative result of our algorithm testing on intent-youtube datasets. Each sequence contains 10 samples uniformed sampled across time. The colorbar depicts intentionality judgement by our algorithm at each frame. Median filter with windows size 30 frames is also applied for visual presentation. }
\label{fig:youtube-result}       
\end{figure*}

The result in last section shows that our algorithm achieves higher or comparable classification accuracy on the video-level to the baseline methods. However, it is unknown if our algorithm can return reasonable segment-level classification. Imagine an agent trips while walking, with walking occupying a significant portion of the video. An algorithm can give correct annotation (non-intentional) of the sequence even if the walking is labeled as non-intentional and tripping as intentional. To examine this possible issue, we provide qualitative result of segment/frame-level intent recognition by our algorithm on the three datasets (see Figure \ref{fig:maya-result} for intent-maya dataset, Figure \ref{fig:mocap-intent} for intentional actions and Figure \ref{fig:mocap-nonintent} for non-intentional actions in intent-mocap dataset, see Figure \ref{fig:youtube-result} for intent-youtube dataset). These results shows that our algorithm can correctly recognizing intent of the agent on both video level and segment level. For example, in the ``Tripping'' sequence, our algorithm correctly annotates that the action (walking) is intentional before the 70th frame and unintentional thereafter, accurately reflecting the moment that the agent trips.

\subsection{Effect of keypoints occlusion}
Since our algorithm depends on the estimation of the agent's center of mass, it is necessary to study the robustness of our algorithm against keypoints occlusion in the pose estimation. 

We design three experiments to simulate a variety of cases of occlusions on skeleton keypoints: 1. A random joint is always occluded across all samples, similar to the cases where one of the sensors (or mocap markers) is defective; 2. A random joint is occluded per agent, which simulate the cases where a keypoint is occluded consistently for an agent; 3. A random joint is occluded per frame, which is to simulate a highly noisy center of mass estimate. Typically, the occlusion on a specific keypoint occurs consecutively across several frames, during which estimation of agent’s center of mass is biased. For our algorithm, a biased but smooth estimate of the center of mass is less problematic than a highly noisy estimate, which might produce an artificial increase of mechanical energy (due to the jittering movement). Thus, we occlude random joints per frame to simulate this highly noisy estimation of the center of mass, which tests our algorithm in a highly unfavorable setup. The keypoint occlusion experiment is only conducted on the intent-mocap and intent-youtube dataset, since the intent-maya dataset does not have keypoints defined for its ball agent. 

Table \ref{tab:keypoint-occlusion} shows the result of our algorithm on intent-mocap and intent-youtube dataset with the three types of keypoint occlusion mentioned above. To provide a measurement of uncertainty, we repeat the three experiments with randomly selected keypoints for occlusion and report the mean accuracy and its standard error. As we can see, when only one joint is consistently occluded, either across all samples or per agent, there is no significant negative impact on our algorithm. In the worse case occlusion we designed for our algorithm, there is a drop in the classification accuracy due to the inaccurate estimation on the change of total mechanical energy induced by the noisy estimation of the center of mass, which is consistent with what we described earlier. Notice that we did not perform preprocessing to smooth the trajectory of the center of mass or infer the missing joint, which can be done to improve the performance. 

\begin{table}[h!]
\centering
\caption{Quantitative result on our algorithm with occluded keypoints, measured by mean classification accuracy and standard error of the mean (in parenthesis). }
\label{tab:keypoint-occlusion}       
\begin{tabular}{l l l}
\hline\noalign{\smallskip}
Occlusion           & intent-mocap    & intent-youtube\\
\noalign{\smallskip}\hline\noalign{\smallskip}
None                & 0.827           & 0.785 \\
1 joint all sample  & 0.833 (0.008)   & 0.769 (0.011)\\
1 joint per agent   & 0.808 (0.006)   & 0.767 (0.003)\\
1 joint per frame   & 0.712 (0.021)   & 0.624 (0.006)\\
\noalign{\smallskip}\hline
\end{tabular}
\end{table}

\subsection{Ablation study}
The result in the previous section shows that our algorithm is effective on recognizing intentionality from the trajectory of the agents. However, it is still unknown that if all the common sense concepts we introduced in the Section \ref{section:method-modeling} are necessary. We conduct an ablation study on the proposed algorithm to study this problem. In this experiment, we started with a model including only Concept 1, and gradually adding each common concept until reaching the full model with all four concepts. When classify with the ablated model, we directly apply the method described in Section \ref{ss:recognition-method} to $I_*$ - the output of the model with ablated Concepts. For example, when only C1 is used, $I_*=I_{C1}$, with $I_{C1}$ defined in equation (\ref{eq:model-C1}). For the ablated model with C1+2+4, we calculate the output of the model using Algorithm \ref{alg:C4} but with $I_{C2}$ as input instead of $I_{C3}$.

\begin{table}[h!]
\caption{Quantitative result of our algorithm with ablation, measured by mean classification accuracy and standard error of the mean (in parenthesis)}
\label{tab:ablation}       
\begin{tabular}{l l l l l l}
\hline\noalign{\smallskip}
dataset & 1 & 1+2 & 1+2+3 & 1+2+4 & 1+2+3+4 \\
\noalign{\smallskip}\hline\noalign{\smallskip}
maya    & 0.500 & 0.667 & 0.667 & 0.783 & 0.950\\
mocap   & 0.500 & 0.571 & 0.534 & 0.737 & 0.827\\
youtube & 0.500    & 0.519    & 0.501    & 0.735    & 0.785\\
\noalign{\smallskip}\hline
\end{tabular}
\end{table}

\subsection{Analysis on the ablation result}
The result of the ablation study is shown in Table \ref{tab:ablation}. When the C1 is the only concept used in the model, the classification accuracy is no greater than random chance for both maya and mocap datasets. With more common sense concepts included, the classification accuracy increases, indicating that the proposed common sense concepts are all necessary to achieve an accurate recognition. One may notice that the accuracy of C1+2+3 is no higher than C1+2 and may argue that C3 is not necessary for this reason. However, this argument is challenged by the result of C1+2+4 is less accurate result than C1+2+3+4, indicating that C3 is necessary in when combined with C4 (rather than C2) to further improve the accuracy. As mentioned in Section \ref{ss:concept-intro}, the four concepts combined to form a common knowledge system for intent recognition. 

\section{Discussion and conclusion}
The result in the previous section shows that our algorithm can achieve significantly superior or comparable accuracy to a range of learning based method on three datasets. The result also shows the necessity of each of the common knowledge concepts in the proposed algorithm. By modeling those concepts that defines intentionality, our algorithm does not need labeled data, nor it is a learning-based algorithm, thus shy away from the potential sampling bias in training set. Since the algorithm does not need training and composes only rules that derived from human common sense, the method is less computationally demanding and more interpretable than most of the deep learning based algorithms. The performance on the three datasets also shows the general applicability of the proposed common concept algorithm on different types of agents. 

The result also shows that the classification accuracy of our algorithm is decreasing from maya to mocap to youtube dataset. One possible explanation is the lack of accuracy in the estimated center of mass of the agent in the mocap and youtube datasets. Intent-maya dataset only contains sphere-like agents, whose trajectory of the center of mass is readily available with high precision. However, in the mocap and youtube datasets, the center of mass has to be estimated from the skeleton of the agent, which is more accurate in mocap dataset than in youtube dataset (but both worse than the maya dataset), potentially explaining the drop of classification accuracy in the yotube dataset comparing to the mocap dataset. One should also notice that the proposed algorithm does not claim nor implement any novelty in 3D human pose estimation, which by itself is a challenging and open problem.  

\begin{table*}[h!]
\centering
\caption{Cases for intentionality of the interaction between A and B.}
\label{table:multi-agent-example}
\begin{tabular}{|c|c|c|c|p{7cm}|}
\hline
Case & A action & B action & A $\rightarrow$ B interaction & Example \\
\hline\hline
1 & intentional & intentional & intentional & A punched B in a boxing game. \\
\hline
2 & intentional & intentional & non-intentional &  A is walking backward but B is walking normally. A bumped into B without noticing B.  \\
\hline
3 & non-intentional & intentional & non-intentional & A is on a bike out of control and crashed into B.  \\
\hline
4 & non-intentional & non-intentional & non-intentional &  A crashed into B when both of them are ice-skiing and cannot control their movement.  \\
\hline
5 & intentional & non-intentional & intentional & A catches B while B tripped over an obstacle.   \\
\hline
\end{tabular}
\end{table*}

Another significant future direction of the study is to infer intentionality when the action involves multiple agents. When social interactions are involved, the inference of intentionality can become much more complex. Table \ref{table:multi-agent-example} provides a rudimentary anecdotal analysis on the potential cases of intentionality in a two-agent system without considering the environmental context. As shown in the table, the relationship is complex, but not lacking of rules. For example, If the action conducted by agent A is non-intentional, it is likely that the interaction from agent A to agent B is also non-intentional. However, the \textit{reverse equality} may not hold. Thus, we argue that to consider solving this complicated multi-agent problem, it is necessary to address the that of a single agent first, which is what we did in the present work.

In conclusion, we proposed a common knowledge based unsupervised computer vision system for recognizing intent of an agent, specifically whether the action of the agent is intentional or not. The problem is significant due to the essential role played by the intent recognition in human's social life. Any machine that intended to work and live with human might benefit from intent recognition to achieve a smooth human-machine interaction. Recognition of intentionality (intentional vs unintentional) is a first step towards this goal. Our algorithm, to our knowledge, provides the first common knowledge based computer vision algorithm for the recognition of intentionality. Comparing to the modern computer vision and pattern recognition systems, whose majority are data-driven learning methods that require a large amount of training data, our system achieves this high-level vision task without the need for training data, but achieves higher or comparable result on multiple datasets to the baselines. The effectiveness of our algorithm not only provides a potential way to address the problem of automatic visual recognition of intent, but also performing high-level reasoning without using training data by leveraging human commonsense concepts.

\begin{acknowledgements}
This research was supported by the National Institutes of Health (NIH), grants R01-DC-014498 and R01-EY-020834, the Human Frontier Science Program (HFSP), grant RGP0036/2016, and a grant from Ohio State's Center for Cognitive and Brain Sciences.
\end{acknowledgements}

\bibliographystyle{spmpsci}      
\bibliography{Intent}   

%
%

\end{document}